\documentclass[review]{elsarticle}

\usepackage{lineno,hyperref}
\usepackage{mathtools}
\usepackage[algo2e]{algorithm2e}
\usepackage{algorithm}
\usepackage{pgfplots}
\usepackage{tabularx}
\usepackage{longtable}
\usepackage{multirow}
\usepackage{textcomp}
\usepackage{verbatim}
\modulolinenumbers[5]
\newcommand{\specialcell}[2][c]{%
	\begin{tabular}[#1]{@{}c@{}}#2\end{tabular}}

\makeatletter
\def\ps@pprintTitle{%
	\let\@oddhead\@empty
	\let\@evenhead\@empty
	\let\@oddfoot\@empty
	\let\@evenfoot\@oddfoot
}
\makeatother

\biboptions{authoryear}

\begin{document}

\begin{frontmatter}

\title{U-CNNpred: A Universal CNN-based Predictor for Stock Markets}

\author[mymainaddress]{Ehsan Hoseinzade}
\ead{ehsan\_hoseinzade@sfu.ca}
\author[mymainaddress2]{Saman Haratizadeh\corref{cor1}}
\ead{haratizadeh@ut.ac.ir}
\author[mymainaddress]{Arash Khoeini}
\ead{akhoeini@sfu.ca}
\cortext[cor1]{Corresponding author}

\address[mymainaddress2]{Faculty of New Sciences and Technologies, University of Tehran, Tehran, Iran}
\address[mymainaddress]{School of Computing Science, Simon Fraser University, Burnaby, Canada}

\begin{abstract}
The performance of financial market prediction systems depends heavily on the quality of features it is using. While researchers have used various techniques for enhancing the stock specific features, less attention has been paid to extracting features that represent general mechanism of financial markets. 
In this paper, we investigate the importance of extracting such general features in stock market prediction domain and show how it can improve the performance of financial market prediction . We present a framework called U-CNNpred, that uses a CNN-based structure. A base model is trained in a specially designed layer-wise training procedure over a pool of historical data from many financial markets, in order to extract the common patterns from different markets. Our experiments, in which we have used hundreds of stocks in S\&P 500 as well as 14 famous indices around the world, show that this model can outperform baseline algorithms when predicting the directional movement of the markets for which it has been trained for. We also show that the base model can be fine-tuned for predicting new markets and achieve a better performance compared to the the state of the art baseline algorithms that focus on constructing market-specific models from scratch.

\end{abstract}

\begin{keyword}
\texttt{Stock markets prediction}\sep Deep learning\sep Convolutional neural networks \sep CNN \sep Layer-wise training \sep Transfer learning
\end{keyword}

\end{frontmatter}


\section{Introduction}
Forecasting financial markets is a tempting challenge for both researchers and investors. It is an intricate task, due to the dynamic, nonlinear and noisy behavior of the markets. In past decades, data mining techniques have been widely used for technical and fundamental analysis in market prediction domain. The recent researches in this field are mostly focused on one of the two main approaches:

First, analyzing news to extract patterns that are somehow correlated to the ups and downs of the stock markets {\citep{vargas2017deep, ding2015deep}}.  Various text mining techniques like bag of words, n-gram, word2vec
{\citep{nassirtoussi2014text, kim2014convolutional}} have been used to convert text data to feature vectors, using which, a classifier can be trained to predict movement direction of stocks. 

The second approach is to predict stock markets by using historical prices. According to efficient market hypothesis, all the information is available in the price. Although this theory found prediction of stock markets impossible 
{\citep{fama1970efficient}},
 recent advances show that’s quite possible. Two well known examples of this category are prediction of continuous value like price, or return and discrete value like direction of movement. Different techniques like artificial neural networks 
{\citep{guresen2011using, enke2005use, kara2011predicting}}
, support vector machine
{\citep{patel2015predicting1, patel2015predicting, hsu2009two, chen2017feature}}
, genetic algorithm
{\citep{atsalakis2009surveying, ahmadi2018new}}
, logistic regression
{\citep{ballings2015evaluating}}
, semi-supervised models {\citep{kia2018hybrid}}
 and random forests 
{\citep{khaidem2016predicting, lohrmann2018classification}}
 have been utilized for this purpose. Most of the time, researchers utilize technical indicators as input features for their prediction models. However these indicators seem to be unnecessarily simple for the prediction task, as they are meant to be used by human market experts. 

Deep learning is a novel field of machine learning, which tries extract more complex and sophisticated features. Usually deep learning algorithms need a lot of data and computational power to get valuable results. Fortunately, by advances in technology, computational power has been improved and the required hardware resources are now available. Thus, after showing successful results is Computer Vision
{\citep{he2016deep}}
and Natural Language Processing (NLP)
{\citep{lecun2015deep}}
, deep learning has became popular in stock market prediction as well.  Algorithms like deep multilayer perceptron (MLP)
{\citep{yong2017stock}}
, restricted Boltzman machine (RBM)
{\citep{cai2012feature, assis2018restricted}}, 
long short-term memory (LSTM)
{\citep{chen2015lstm}}
, autoencoder (AE)
{\citep{bao2017deep}} 
are famous deep learning algorithms utilized to achieve better predictions. To the best of our knowledge, applications of CNN in stock market prediction have been mentioned in few papers  including 
{\citep{gunduz2017intraday}}
 . However, CNN is renowned for its feature extraction abilities that have been demonstrated in other domains like Computer Vision, Natural Language Processing and Speech Processing.
 
Financial markets have some common general characteristics. Actually it is the reason for the fact that the technical indicators provide useful information for prediction of many markets throughout the world. This general similarity between different markets makes it possible to find common features that can be applied by prediction systems to forecast the future behavior of several markets. Examples of this can be found in {\citep{hoseinzade2019cnnpred, gunduz2017intraday, chong2017deep}}. However, such prediction systems use deep learning methods that  need to be trained using a big repository of historical data from every market that they are going to predict. That means that the patterns extracted by the prediction model is so much adapted to the observed training data, that can not be easily applied for predicting new markets. In other words,the existing approaches fail to extract the general dynamics of stock markets; a problem that can reduce the power of models in forecasting the behavior of new markets.   

The general class of learning approaches that focus on handling this issue is called transfer learning. Transfer learning has been successfully applied in the other fields like Computer Vision and Natural Language Processing {\citep{pan2010survey, shin2016deep}}. A question that comes to mind could be “is it possible to use available data for creating a universal predictor that would be able to directly its knowledge for prediction of new stocks too?”. A short answer to this question is probably "No", since different stock markets have different characteristics, including their stocks, their locality, and their size that can certainly affect their dynamics. So a model trained for a market will probably fail to predict a new market if that market is not introduced to the model at all. However, an intermediate solution is  to develop models that can extract the general patterns in their initial training phase and then "tune" themselves for predicting new markets by a secondary training phase. 

In this paper we investigate this subject, to see if the general similarity between markets allows an appropriately designed and trained deep model to be a universal predictor, that is it can predict the future of stocks that it has been trained for, as well as new markets, possibly after a phase of tuning.
 
The key point in our approach is that preventing the training phase from getting too much adapted to the observed data, probably helps the model to extract essential general patterns instead of market-dependent ones. This is especially important when using deep learning methods, since they are prone to overfitting that leads to undesirable convergence 
 {\citep{larochelle2009exploring, hinton2006fast, bengio2007greedy}}
 Inspired by unsupervised initial weighting methods, we suggest to use a layer-wised approach that uses the initial training data to extract the general patterns and leave the details for the later tuning steps. 
 
 We choose CNN as the basis for our framework, U-CNNpred which stands for Universal CNNpred, due to its ability in feature extraction.
 U-CNNpred is capable of predicting new stocks by fine tuning the base predictor for that stock. To test the performance of U-CNNpred in a daily directional movement prediction setting, a data set of 458 stocks in S\&P 500 are fed to the network. In addition, five major indices of the U.S., S\&P 500, NASDAQ, Dow Jones Industrial average, NYSE, RUSSELL and nine famous indices around the world, DAX, Nikkei, KOSPI, FCHI, FTSE, HSI, SSEC, BSESN and NSEI are used to test the performance of U-CNNpred in prediction of new stocks.
 
The main contributions of this work can be condensed as follows:
 
\begin{itemize}
    \item introducing a stock prediction method that outperforms other state of the art algorithms in terms of prediction accuracy and the generalization power.
	\item Introducing a method for transfer learning for stock markets prediction domain. 
	\item Applying a layer-wise approach for training a deep convolutional neural network that successfully extracts general patterns and avoids converging to market-dependent optimums.
\end{itemize}

The rest of this paper are organized as follows. In the next section, related works and researches are presented. Then, we introduce a brief explanation of used algorithm in this research. Next, we describe our CNN and universal predictor in details. After that data description and experimental results are presented. Finally, we summarize our paper.

\section{Related works}

In this section, we briefly review the related works. ANNs that have been used in stock market prediction can be divided into two groups of Shallow ANNs and deep ANNs.

Shallow ANNs have been used in stock prediction for a long time and they are mostly limited to shallow feedforward ANNs. According to many surveys {\citep{krollner2010financial}}, they are the most popular tools among all of the traditional machine learning algorithms that have been used in this domain {\citep{{dai2012combining, ticknor2013bayesian, de2013applying, guresen2011using}}}. The main characteristic of them is one hidden layer which makes them completely different from deep models and also provides them a limited power in feature extraction. In {\citep{kara2011predicting}}, researchers successfully used a shallow ANN to predict directional movement of Istanbul Stock Exchange (ISE) National 100 Index. Features used for prediction were ten technical indicators of the corresponding market. They also have shown the superiority of shallow ANN over an SVM method in their work. Back-propagation is the prevalent algorithm that have been used for training ANNs {\citep{hecht1992theory, hagan1994training}}. 

Usually, due to the noisy and nonlinear behavior of stock markets, the typical back-propagation process in neural networks is susceptible to converging to local optimum. Some researchers have used search techniques like GA and SA to facilitate the learning process {\citep{kim2000genetic, qiu2016application, qiu2016predicting}}. In {\citep{qiu2016application}}, authors find optimal primary weights for network using a search technique, while in another research {\citep{qiu2016predicting}}, final weights of an ANN are directly set by genetic algorithm. 

Incorporating feature extraction algorithms like PCA for achieving better features was the other idea used by researchers to enhance the accuracy of prediction {\citep{zhong2017comprehensive, zahedi2015application}}. In {\citep{zhong2017forecasting}}, various sources of data were passed to a shallow ANN after extracting features using PCA and two other variations of that. The target market was S\&P 500 and results showed up to 59\% accuracy in directional movement prediction. 

Deep learning methods which are renowned for their feature extraction power, have recently been used in this domain as well.
Deep feedforward ANNs, which have more than one hidden layer are one of the first deep learning algorithms used for stock prediction{\citep{chatzis2018forecasting}}. 

In {\citep{moghaddam2016stock}}, both shallow and deep structure of feedforward ANNs were examined in order to achieve a more accurate prediction for NASDAQ. Price of four and nine days before the prediction day were fed to different neural networks and the results revealed that the deep structures outperform the shallow ones. Another example of deep ANN utilization can be found at {\citep{arevalo2016high}}, in which the price of Apple Inc. stock were predicted. Three features were extracted from raw data and passed to a feedforward ANN with five hidden layers for prediction. Reported results showed directional accuracy of 63\% to 65\% during the financial crisis of 2009. In {\citep{yong2017stock}}, researchers have successfully predicted the index price of Singapore’s stock market using a feedforward ANN with three hidden layers. The reported MAE of next day prediction was 0.75 for the suggested deep method that is lower than the prediction errors of its competitor algorithms, that are reportedly more than 1. 

In {\citep{chong2017deep}}, a comparison between PCA, Autoencoder and RBM with respect to their feature extraction capabilities has been done. After extracting new features from 380 initial features with mentioned methods, they were passed to a feedforward ANN with two hidden layers. The results of prediction, showed that none of the methods dominates the others in all of the tests that were done. 

Another kind of neural networks that has been used in this domain are the recurrent neural networks. Their internal memory which is useful for prediction of sequential data has enticed researchers to apply them for stock markets prediction. {\citep{rather2015recurrent, shen2018deep, balaji2018applicability}}. Among different types of RNNs, LSTM seems to be more popular than others {\citep{sang2018improving, chung2018genetic, baek2018modaugnet, sagheer2019time, kim2018forecasting, lahmiri2019cryptocurrency}}. In {\citep{nelson2017stock}}, five stocks in Brazilian stock market were predicted using LSTM. Many technical indicators generated form historical data were fed to an LSTM in order to forecast directional movements of markets. The best reported accuracy is 55.9\% which shows an improvement over the baseline algorithms, including random forest and Multi-Layer Perceptron.

In {\citep{fischer2018deep}}, directional accuracy of the constituent stocks of S\&P 500 were predicted using LSTM. The reported results revealed the superiority of LSTM over the memory-less algorithms like deep neural networks and logistic regression. The overall accuracy of prediction is reported to be about 51\% to 54\%.

The more recently used deep models in this domain are Convolutional Neural Networks {\citep{tsantekidis2017forecasting, nino2018cnn}}. In {\citep{di2016artificial}}, S\&P 500 index was predicted using three different neural networks which were MLP, LSTM and CNN. Input data was only the prices of S\&P and there were no other features. The reported accuracy for CNN is  53.6\%, that is higher than that of MLP and LSTM. An improvement of accuracy has been achieved by combining all of the mentioned methods, that has loed to accuracy of about 57\%. Another example of using CNN in stock market prediction is {\citep{gunduz2017intraday}}. Historical data of 100 stocks in Borsa Istanbul has been used to extract technical features which were then fed to a deep CNN. The authors have shown that clustering the features and arranging the CNN input so that the similar features are close to each other improves the quality of prediction.

In {\citep{hoseinzade2019cnnpred}}, researchers have predicted five major indices using the several data sources including technical indicators, exchange rate of currencies, T-bill data, return of world indices, big U.S. companies, future contracts, and price of commodities. These features were utilized by two variations of a CNN based framework, called CNNpred. While 2D-CNNpred's input were two-dimensional tensor and a general model were trained for forecasting all the markets, 3D-CNNpred utilized three-dimensional input and each index were predicted using a separated CNN. Reported results showed that both of these methods outperformed baseline algorithms and achieve Macro Averaged F-measures in the range of 47\% to 49\%. 

In  {\citep{sezer2018algorithmic}} researchers have applied CNN on different stocks and ETFs in U.S. market data to predict trading signals. Input data is a $15\times 15$ matrix consisted of the history of the last 15 days of 15 technical indicators. The goal of this research was to learn a model to decide among buy, sell or or hold, for each day, in order to maximize the overall return. The paper reports the superiority of CNN over other baseline algorithms with an average annual return of about 13\%.

All the mentioned papers are summarized and compared with regards to their training and prediction types in Table \ref{table:summary}. While in all of the mentioned papers predictions are made using either a customized model for each market or a global model trained by all of the markets data, none of the presented algorithms are able to predict new markets for which they have not been directly trained. Our proposed framework is capable of extracting general features of the markets in the data set and then use them to predict new markets by fine-tuning the trained network. Meanwhile we aim to address the issue of converging to local optimum by utilizing a layer-wise training procedure in our suggested approach.

\begin{table}
	\begin{center}
		\footnotesize
		\resizebox{1\textwidth}{!}
		{
			\begin{tabular}{c c c c c }
				\hline
				Author/year & Target Data & \specialcell{Training \\Layers} & \specialcell{Prediction \\Type}  &  \specialcell{Prediction\\ Method}  \\ 
				\hline \hline
				{\citep{kara2011predicting}} & \specialcell{Borsa Istanbul \\BIST 100 Index} & Simultaneously & \specialcell{Customized} & \specialcell{ANN\\ SVM} \\\\
				{\citep{qiu2016application}} & \specialcell{Nikkei 225 \\index} & Simultaneously & Customized & \specialcell{GA+ANN \\SA+ANN} \\\\
				{\citep{qiu2016predicting}}& \specialcell{Nikkei 225 \\index} & Simultaneously & Customized & GA+ANN \\\\
				{\citep{nelson2017stock}}& \specialcell{Brazil Bovespa\\5 stocks} & Simultaneously & Universal & LSTM \\\\
				{\citep{di2016artificial}}& \specialcell{S\&P 500 index} & Simultaneously & \specialcell{Customized} & \specialcell{MLP RNN\\CNN} \\\\
				{\citep{moghaddam2016stock}}& \specialcell{NASDAQ } & Simultaneously  & Customized & ANN-DNN  \\\\
				{\citep{yong2017stock}}& \specialcell{Singapore STI } & Simultaneously  & Customized & DNN  \\\\
				{\citep{arevalo2016high}}& \specialcell{AAPL Inc. } & Simultaneously  & Customized & DNN \\\\
				{\citep{zhong2017forecasting}}& \specialcell{S\&P 500 index } & Simultaneously  & Customized & ANN  \\\\
				{\citep{chong2017deep}}& \specialcell{Korea KOSPI \\38 stock returns} & Simultaneously & Universal & \specialcell{PCA-RBM \\ AE + DNN} \\\\
				{\citep{fischer2018deep}}& \specialcell{S\&P Stocks} & Simultaneously & Universal & LSTM \\\\
				{\citep{gunduz2017intraday}}& \specialcell{Borsa Istanbul \\BIST 100 stocks} & Simultaneously & Universal & CNN \\\\
				{\citep{hoseinzade2019cnnpred}} & \specialcell{U.S. 5 \\major indices} & Simultaneously & Universal & CNN \\\\
				{\citep{sezer2018algorithmic}}& \specialcell{U.S. ETFs \\and stocks} & Simultaneously & Universal & CNN \\\\
				Our method& \specialcell{U.S. stocks \\ global indices} & layer-wise & \specialcell{Universal\\ transfer learning} & CNN \\\\
		
			\end{tabular}
			}
		\caption{Summary of explained papers}
		\label{table:summary}
		
	\end{center}
\end{table}
\section{Background}
In this section, we briefly introduce the concepts and algorithms used in our framework including convolutional neural networks, how they are trained and the concept of transfer learning.

\subsection{Convolutional neural network}
Convolutional neural networks were inspired by human visualization system. LeCun and some of his colleagues invented them for the first time in 1995  \citep{lecun1995convolutional,gardner1998artificial}. Each CNN like other neural networks consists of input, output and hidden layers. Most of the time, hidden layers comprise three main layers of convolutional, pooling and fully connected layers. This neural network belongs to the category of supervised learning algorithms in which a specific target is associated with each sample data.

\subsubsection{Convolutional layer}
Convolutional layer performs convolutional operation by defining filters with certain size. Size of a filter reflects the area of data it covers at one step. Inside a filter, there are numbers which indicate weights of convolutional operation and they are updated during the process of training. Outputs of the convolutional operation are passed through an activation function, such as sigmoid, tanh or Relu \ref{eq:relu}, that has been widely used in the literature of CNN for two reasons of nonlinearity and efficiency.

\begin{equation}
	f(x) = max(0,x)
	\label{eq:relu}
\end{equation}

By considering input of layer $l-1$, an $N\times N$ matrix and convolutional filters $F\times F$, input of layer $l$ is calculated according to Eq \ref{eq:conv} and Eq \ref{eq:conv_activation}. 

\begin{equation}
	x_{i,j}^{l} = \sum_{k=0}^{F-1}\sum_{m=0}^{F-1}w_{m,k}v_{i+k,j+m}^{l-1}	
	\label{eq:conv}
\end{equation}

\begin{equation}
	v_{i,j}^{l} = \sigma(x_{i,j}^{l})
	\label{eq:conv_activation}
\end{equation}

In the above formula, $v^{l}_{i,j}$ is the value at row $i$ and column $j$ of layer $l$, $w_{k,m}$ is the weight at row $k$ and column $m$ of filter and $\delta$ represents the activation function.

\subsubsection{Pooling layer}
Pooling layers are usually after convolutional layers and they are supposed to perform a subsampling task. Reducing computational costs is one of the main incentives behind the pooling layers, because it reduces size of data at each step. In addition, they can be seen as one the CNN\textquotesingle s modules that controls overfitting. Beside that, they make extracted features invariant to transformation, since the exact location of extracted features is not important after that. Max pooling and Average pooling are two kinds of pooling. In Max pooling, maximum value inside a pooling window is chosen for passing to the next layer. 

\subsubsection{Fully connected layer}
At the end of each CNN, there is a feedforward neural network which takes the output of convolutional and pooling layers and generates the final result of the whole network. In fact, convolutional and pooling layers extract local features while the fully connected layer relates those features to the final result. Eqs \ref{eq:fully_connected} and \ref{eq:activation} illustrate  how a fully connected layer generates outputs.

\begin{equation}
	x_{i}^{l} = \sum_k v_{k}^{l-1}w_{k,i}^{l-1}
	\label{eq:fully_connected}
\end{equation}

\begin{equation}
	v_{i}^{l} = \sigma(x_{i}^{l})
	\label{eq:activation}
\end{equation}

In Eq \ref{eq:activation}, $v_i^l$ is the value of neuron $i$ at the layer $l$, $\delta$ is activation function and weight of connection between neuron $k$ from layer $l-1$ and neuron $i$ from layer $l$ are shown by $w_{k,i}$. 

\subsection{Training a CNN}
Training a CNN happens in three steps of forward propagation, backward propagation and updating the weights of the network. We here briefly review these three steps.

\subsubsection{Forward Propagation}
This phase starts with feeding the input to the input layer. Then, convolutional and pooling layers are applied to the input. Output of these layers are passed to fully connected layers to generate the final results.

\subsubsection{Backward Propagation}
Because of the difference between the Convolutional layer and the fully connected layer, this step is different for each of these two layers.

First, error is calculated at the output layer through Eq \ref{eq:back_full1}. Using Eq \ref{eq:back_full2} and Eq \ref{eq:back_full3}, error is calculated at the fully connected layer. The gradient for the corresponding weights is calculated using Eq \ref{eq:back_full4}.

\begin{equation}
	\frac{\partial E}{\partial v_i^L} = \frac{d}{d v_i^L} E(v^L) 
	\label{eq:back_full1}
\end{equation}

\begin{equation}
	\frac{\partial E}{\partial x_j^l} = \sigma^{\prime}(x_j^l) \frac{\partial E}{\partial v_j^L} 
	\label{eq:back_full2}
\end{equation}

\begin{equation}
	\frac{\partial E}{\partial v_j^l} = \sum w_{i,j}^{l} \frac{\partial E}{\partial x_j^{l+1}} 
	\label{eq:back_full3}
\end{equation}

\begin{equation}
	\frac{\partial E}{\partial w_{i,j}^{l}} = v_i^l \frac{\partial E}{\partial x_j^{l+1}} 
	\label{eq:back_full4}
\end{equation}

Same happens in convolutional layers. However, because of the way convolutional layers operate, there are some changes in the formula (Eqs \ref{eq:conv2} and \ref{eq:conv3}). Our final goal is to compute the gradient of weights of filters of the convolutional layers. This happens through Eq \ref{eq:conv1}.

\begin{equation}
	\frac{\partial E}{\partial w_{i,j}^{l}} = \sum_{k=0}^{N-F}\sum_{m=0}^{N-F} \frac{\partial E} {\partial x_{m,k}^{l}} v_{i+k,j+m}^{l-1} 
	\label{eq:conv1}
\end{equation}

\begin{equation}
	\frac{\partial E}{\partial x_{i,j}^{l}} =  \frac{\partial E} {\partial v_{i,j}^{l}} \sigma^{\prime}(x_{i,j}^{l}) 
	\label{eq:conv2}
\end{equation}

\begin{equation}
	\frac{\partial E}{\partial v_{i,j}^{l-1}} = \sum_{k=0}^{F-1}\sum_{m=0}^{F-1} \frac{\partial E} {\partial x_{i-k,j-m}^{l}} w_{k,m} 
	\label{eq:conv3}
\end{equation}

\subsubsection{Updating the weights}
The last step is updating weights of the CNN using calculated gradient in the backward propagation. Researchers have proposed many optimization algorithms to achieve this aim. Stochastic Gradient Descent {\citep{bottou2010large, robbins1951stochastic}} Adam {\citep{kingma2014adam}}, Adagrad {\citep{duchi2011adaptive}} and Adadelta {\citep{zeiler2012adadelta}} are among methods that are applied in for this step.

\subsection{Transfer learning}
In other applications of deep learning like Computer Vision and Natural Language Processing, too much data may be needed for training a deep neural network from scratch and repeating the time consuming process of training for every similar dataset may seem unreasonable. As a solution, researchers can use a method called Transfer Learning {\citep{pan2010survey, shin2016deep}}. In this approach, a model is trained using an initial data set available for a certain problem in a way that, when a new sufficiently-similar problem arises and a new model is needed for it, the previously-trained model can be tuned for it possibly using a new set of data. This approach can be useful not only to save time, but also to achieve better results, as it allows the model to extract the general knowledge that is useful for solving a class of problems, and use it to generate finely tuned solutions for each problem in that class. 

\section{Proposed framework: U-CNNpred}
As we mentioned earlier, financial markets have general similar characteristics. Transfer learning is also a successful and renowned process of transferring the obtained knowledge from similar domains like various financial markets. The performance of the whole process heavily depends on the performance of the base model. To get the maximum performance of the base model and avoiding convergence to undesirable local optimums, we propose using a layer-wise method for training a CNN-based model. We call the resulting base model the base predictor, that is then fine-tuned before it can be used as a prediction model in a certain new market. This fine-tuning step could happen in different ways, depending on whether it affects the whole network of the universal predictor, or only a part of it. In this paper we will inspect three versions of the U-CNNpred:

\begin{itemize}
	\item Base Predictor: In which prediction is made by the base predictor, that is a CNN structure trained in a layer-wise manner.
	\item partially fine-tuned Predictor: In which prediction for each market is done by the base predictor, after fine-tuning its last layer for that market.
	\item complete fine-tuned Predictor: in which prediction for a new market is done by the base predictor after fine-tuning all of its weights for that market.
\end{itemize}
Algorithm \ref{algo: U-CNNpred} describes various steps of U-CNNpred. Mentioned steps are described more specifically in the next sections.

\subsection{Base CNN}
CNN is renowned for its ability in feature extraction, however the quality of prediction depends on some parameters including the number of hidden layers, structure of them, dropout rate, number of filters in each layer, size of each filter and so on. Thus, it is important to use CNN with structure and parameters that fits the problem. 

\begin{algorithm}[H]
	\scriptsize
	\SetAlgoLined
	\KwData{stock, new-stock, history}
	\KwResult{results, half-results, full-results }
	base CNN = define structure of CNN \\
	representation = []\\
	\textbf{//}Combine data of all the markets\\
	\For{$i=1$ to \# of stocks}{ 
		Add data of stock[i] to representation 
	}
	base predictor = layer-wise training (base CNN, representation)\\
	\textbf{//}Prediction of test section of stocks used in training\\
	results = predict(base predictor, representation)\\
	\textbf{//}Market Specific Tuning\\
	\For{$i=1$ to \# of new stocks}{ 
		partial-model = partial fine-tuning(base predictor, new stocks[i])\\
		Add predict(partial-model, res-stock) to partial-results\\
		full-model = complete fine-tuning(base predictor, rep-stock)\\
		Add predict(full-model, res-stock) to full-results
	}
	return results, partial-results, full-results
	
	\caption{U-CNNpred}
	\label{algo: U-CNNpred}
\end{algorithm}

\subsection{layer-wise training}
Traditional algorithms of training a CNN have used unsupervised learning methods to give better initial weights to the network. However, CNN is a supervised algorithm and it seems promising to utilize a methods which finds initial weights according to label of data rather than ignoring it. So, we break base CNN into different subCNNs and incrementally grow them towards the desired CNN structure, by train each subCNN using weights of a smaller previously trained subCNN. In other words, final weights of previously trained subCNNs are used as initial weights of the next subCNN, and this process continues until the CNN structure is complete. This approach relies on the fact that smaller subCNNs have fewer weights to be learnt and are less prune to overfitting. On the other hand, such a subCNN is not flexible enough to learn all the details. So we train a smaller subCNN to grab the general patterns required for prediction and then expand it to make the learned patterns more accurate.
 
Let’s suppose our network consists of an input layer, $n$ trainable hidden layers (convolutional and fully connected, not pooling layers) and output layer. The training phase is done in $n-1$ steps. At step $i$, the network includes input layer, all the layers up to the $i+1^{th}$ trainable layer and output layer. The trained subCNN at step $i-1$ provides the initial weights of the similar layers at step $i$. This process stops when all the $n-1$ steps are done and the last trained subCNN that its structure matches the base CNN, is our final model. Algorithm \ref{algo: layer-wise} describes what mentioned here. Output of layer-wise training is called the base predictor.

\begin{algorithm}[H]
	\scriptsize
	\SetAlgoLined
	\KwData{base CNN, data}
	\KwResult{new-model}
	split data into train, validation, test \\
	\textbf{//}save current model in each step to use it's weights for initial weights of next model \\
	prev-model = new-model = []\\
	\For{$i=1$ to \# of layers}{ 
		new-model = base CNN[input:$i$] + Output layer \\
		\If{$i >$ 1}{
			initial weights new-model[input:$i-1$] = weights prev-model[input:$i$] 
		}
		initial weights new-model[$i-1$:output] = Randomly\\
		new-model = Train (new-model, train, validation)\\
		prev-mode = new-model
	}
	return new-model
	\caption{layer-wise Training}
	\label{algo: layer-wise}
\end{algorithm}

\subsection{Transfer learning}
In order to measure stability of universal predictor, it is used for prediction of new stocks. Hopefully, since base predictor is trained by data of many stocks, it captures fundamental features that can represent the general behavior of the price in stock markets. However, since new stocks may have their own specific behavioral details, it seems promising to fine-tune the base predictor for new stocks. As we mentioned before we this could happen in two ways of partial fine-tuning and full fine-tuning. 

\begin{itemize}
	\item \textbf{partial fine-tuning:} In this approach, Only the weights of the last layer of the base predictor are updated and fine-tuned for the new market while the other layers do not change. This approach is based on the assumption that the extracted features in the base predictor can be used to model the dynamics of new stock markets but to do so, the final layer may need to be fine-tuned so that the model can represent the markets behavior in terms of those extracted fundamental features.
	\item \textbf{complete fine-tuning:} Another possible option for making the base predictor compatible with new markets is to fine tune the whole base predictor by historical data of that market. This approach, unlike the first one, assumes that the extracted features may also need to be further tuned and adjusted before they can be applied for modeling the behavior of a new market.
\end{itemize} 

\section{Experimental settings and results}
In this section, results of our experiments are  presented and discussed. Before that, we introduce the dataset, parameters of U-CNNpred, evaluation methodology and baseline algorithms. 

\subsection{Dataset description}
To test the performance of U-CNNpred, we use two different datasets. The first one contains the daily close prices of 458 stocks in S\&P 500 index, including XOM, JPM, AAPL, MSFT, GE, JNJ, WFC and AMZN. These stocks are used to train the base predictor. In addition, we use data of S\&P 500 index, NASDAQ Composite, Dow Jones Industrial Average, NYSE Composite, RUSSELL 2000, DAX, KOSPI, Nikkei, CAC, FTSE, HSI, SSE Composite, BSE 30 and NIFTI 50 to test the performance of U-CNNpred in forecasting new stocks. More information about mentioned stocks and indices is available at Table \ref{table:dataset}. Every sample has 82 features that can be categorized into eight groups of primitive, technical indicators, big U.S. companies, commodities, exchange rate of currencies, future contracts and world’s stock indices, and other sources of information {\citep{hoseinzade2019cnnpred}}. More information about them is available in Appendix I. Eq \ref{eq: target} shows how label of each sample is calculated.

\begin{table}
	\centering
	\scriptsize
	\begin{tabular}{c| c }
		Name & Description  \\ 
		\hline \hline
		 S\&P 500 & Index of 505 companies exist in S\&P stock market \\
		 Dow Jones Industrial Average & Index of 30 major U.S. companies \\  
		 NASDAQ Composite & Index of common companies exist in NASDAQ stock market \\
		 NYSE Composite & Index of common companies exist in New York Stock Exchange\\
		 RUSSEL 2000 &  Index of 2000 small companies in U.S. \\
		 DAX &  Index of 30 major German companies \\
		 KOSPI & Korea Stock Exchange index  \\
		 Nikkei & Stock market index for Tokyo Stock Exchange  \\
		 CAC & French stock market index  \\
		 FTSE & Index of 100 companies exist in London Stock Exchange  \\
		 HSI & Hong Kong stock market index  \\
		 SSE Composite & Shanghai Stock Exchange index  \\
		 BSE 30 & Index of 30 companies exist in Bombay Stock Exchange  \\
		 NIFTI 50 & Index of 50 companies exist in National Stock Exchange of India \\
		 XOM & Exon Mobil Corporation \\ 
		 JPM & JPMorgan Chase \& Co. \\
		 AAPL & Apple Inc. \\
		 MSFT & Microsoft Corporation \\
		 GE & General Electric Company \\
		 JNJ & Johnson \& Johnson \\
		 WFC & Wells Fargo \& Company \\
		 AMZN & Amazon.com Inc. \\	
	\end{tabular}
	\caption{Description of used indices}
	\label{table:dataset}
\end{table}

\begin{equation}
	target = 
	\begin{cases}
	1 & Close_{t+1}>Close_t \\
	0 & \text{else }\\
	\end{cases}
	\label{eq: target}
\end{equation}

Where $Close_{t}$ refers to the closing price at day t.

These data are related to the period of Jan 2010 to Nov 2017. All the time series in our dataset are split into three categories of train, validation and test. In the first dataset, 458 stocks of S\&P 500, data of Jan 2010 to Jul 2015 are used as train and validation data. Data of Jul 2015 to Apr 2016 are used as test data. Same applies for the train and validation data of the second dataset, 14 famous indices around the world. However, to test the robustness of U-CNNpred through time, when there is a time gap between training data and test data, we define test data to belong to the period of Apr 2016 to Nov 2017.

Since the range of different features may be considerably different, we normalize them using Eq \ref{eq: normalization}, where $x_{new}$ is the normalized feature vector, $x_{old}$ is the original feature vector, $\bar{x}$ and $\sigma$ are the mean and the standard deviation of the original feature vectors.

\begin{equation}
	x_{new} = \frac{x_{old} - \bar{x}}{\sigma}
	\label{eq: normalization}
\end{equation}

\subsection{A base CNN for our dataset}
We use data from 458 different stocks as well as famous indices around the world to test the performance of U-CNNpred. History of each sample includes 60 days and each day consists of 82 features. Consequently, the input to the base CNN is a matrix of 60 by 82. We use 2D-CNNpred as base CNN {\citep{hoseinzade2019cnnpred}}. 

\begin{figure}[!h]
	\makebox[\textwidth]{\includegraphics[width=0.8\paperwidth]{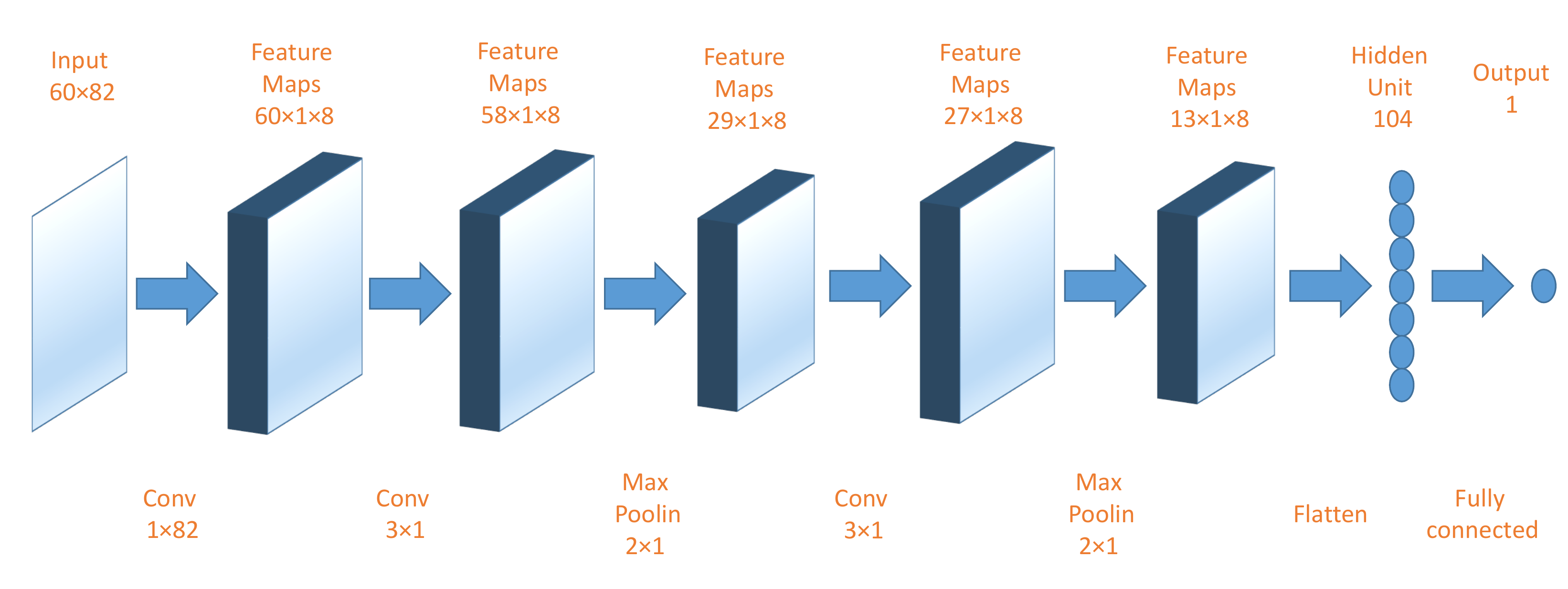}}
	\caption{Graphical Visualization of 2D-CNNpred {\citep{hoseinzade2019cnnpred}}}
	\label{fig: 2D-CNNpred}
\end{figure}

The first convolutional layer in 2D-CNNpred uses eight filters to perform $1\times 82$ (number of features) convolutional operation, in which daily features are combined and produce eight new features. This layer has the potential to act as a feature selection module as well. 
The next step is to combine new daily extracted features to extract more sophisticated ones. The convolutional layers are defined to be $3\times 1$ in order to cover three consecutive days. After the first convolutional layer, two $3\times 1$ convolutional layers  are put in which each of them is pursued by a layer of $2\times 1$ max-pooling layer. Finally, 104 generated features are converted to one final output through a fully connected layer. Fig \ref{fig: 2D-CNNpred} shows a graphical visualization of what we described just now.

\subsection{Evaluation methodology}
Accuracy is the most common evaluation methodology that is used in the classification domain. However, it may not be the best choice for applications in which the data is imbalanced, as it may fail to reflect the true performance of the classification by by being biased towards predictors that prefer the most common class to others. As a solution, here we use the Macro-Averaged-F-Measure to evaluate the performance of the models. This metric reports the mean of F-measure values for each of the two classes of Up and Down {\citep{gunduz2017intraday, ozgur2005text, hoseinzade2019cnnpred}}. 
  
\subsection{Baseline algorithms}
To show the performance of U-CNNpred, we compare it with the following baseline algorithms:

\begin{itemize}
	\item The first baseline algorithm is a combination of PCA and shallow ANN. First, PCA is applied to the input data to extract better features. Then, new features are used by a shallow ANN for prediction {\citep{zhong2017forecasting}}.
	\item The second baseline algorithm utilizes ten technical indicators to train a shallow ANN for forecasting stock markets {\citep{kara2011predicting}}.
	\item The third baseline algorithm is a CNN {\citep{gunduz2017intraday}}. First, by clustering features into different groups, input data is reordered in a way that similar features are put beside each other. Then, a CNN with certain structure is applied to input data.
	\item The fourth baseline algorithm is also a CNN-based one {\citep{hoseinzade2019cnnpred}}, 2D-CNNpred. 2D-CNNpred uses a 2-dimensional tensor to represent input data and uses a CNN structure with traditional training process for building the model. This is the most similar method to the one presented in this paper, as we here adopt the CNN structure used in 2D-CNNpred and apply our suggested algorithm of learning to it.
\end{itemize}

\subsection{Results}
Results of three different experiments are reported in this section. 

\begin{itemize}
	\item Comparing our framework with baseline algorithms.
	\item Measuring how making the base predictor deeper could affect the quality of prediction.
	\item Checking the robustness of U-CNNpred in prediction of new markets in presence of time gap between training data and test data.
\end{itemize}

 One of the baseline algorithms uses PCA and because of that we test this algorithm with different principal components. Consequently, all the other algorithms are tested multiple time. Reported results are average of execution of algorithms for several times. More information about used notation is mentioned at Table \ref{table:algorithms}. 

\begin{table}
	\centering
	\scriptsize
	\begin{tabular}{c| c}	
		Algorithm & Explanation \\ 
		\hline \hline
		B pred & base predictor (Our model) \\
		C Pred & complete fine-tuning (Our model) \\
		P Pred & partial fine-tuning (Our model) \\
		2D-CNNpred {\citep{hoseinzade2019cnnpred}} & base CNN \\
		CNN-cor {\citep{gunduz2017intraday}}& Clustering and reordering feature, CNN as feature extractor and classifier\\
		PCA+ANN{\citep{zhong2017forecasting}} & PCA as dimension reduction and ANN as classifier  \\ 
		Tech+ANN{\citep{kara2011predicting}} & Technical indicators and ANN as classifier  \\
		
	\end{tabular}
	\caption{Description of used algorithms}
	\label{table:algorithms}
\end{table}

\begin{figure}[!h]
	\makebox[\textwidth]{\includegraphics[width=0.8\paperwidth]{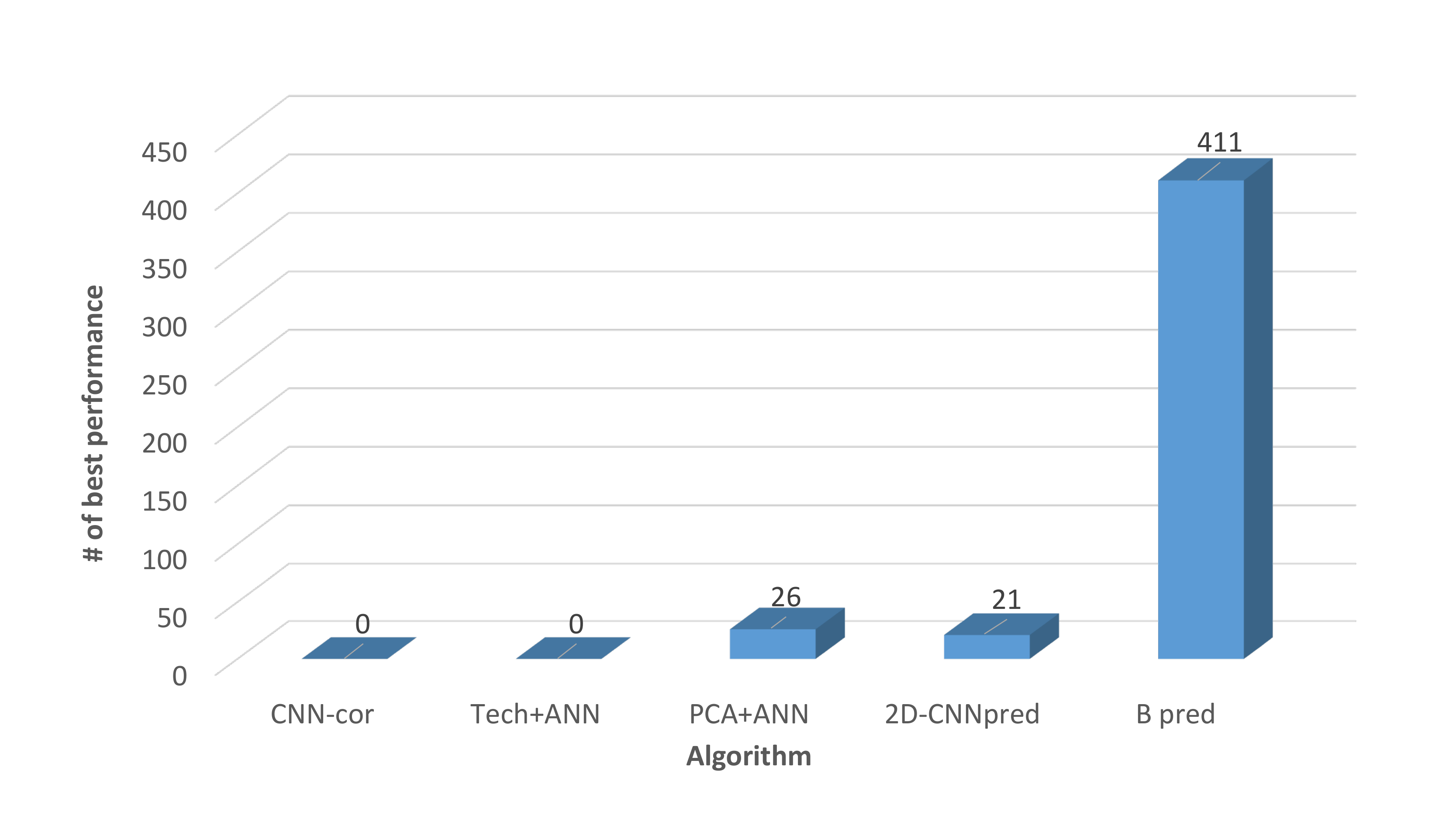}}
	\caption{Performance of algorithms in having the best F-measure for predicting each of the 458 stocks}
	\label{fig: results}
\end{figure}

First experiment consists of feeding 458 stocks to the baseline algorithms as well as our models. The average prediction's F-measure for the whole dataset is reported in Table \ref{table:acccompanies} . To make it more sensible, we also report the prediction's F-measure for 8 big U.S. stocks {\citep{zhong2017forecasting}}. In addition, we counted in how many stocks, out of 458, each algorithm showed the best performance compared to the other algorithms. Fig \ref{fig: results} summarizes the results.

\begin{table}
	\centering
	\scriptsize
	\begin{tabular}{c| c| c | c | c | c}
		Target stock \textbackslash Model & Tech+ANN & PCA+ANN & CNN-cor & 2D-CNNpred  & B pred\\ 
		\hline \hline
		Whole data & 0.4425 & 0.4501 & 0.3569 & 0.4942 & \textbf{0.5111} \\
		XOM & 0.4266 & 0.5032 & 0.3567 & 0.4969 & \textbf{0.5109} \\ 
		JPM & 0.4408 & 0.4626 & 0.3689 & 0.4936 & \textbf{0.5176} \\
		AAPL & 0.4615 & 0.4517 & 0.3670 & 0.4889 & \textbf{0.5058} \\
		MSFT & 0.4507 & 0.4667 & 0.3416 & 0.4961 & \textbf{0.5213} \\
		GE & 0.4077 & 0.4588 & 0.3462 & 0.4973 & 0.\textbf{5161} \\
		JNJ & 0.4290 & 0.4625 & 0.3641 & 0.4840 & \textbf{0.5071} \\
		WFC & 0.4151 & 0.4395 & 0.3637 & 0.4949 & \textbf{0.5102} \\
		AMZN & 0.4993 & 0.4376 & 0.3531 & 0.4962 & \textbf{0.5114} \\
	\end{tabular}
	\caption{F-measure of some of the stocks in S\&P 500 index}
	\label{table:acccompanies}
\end{table}

The second experiment is performed to measure the effect of making the network deeper. While training the base predictor in a level-wise manner, layers are added step by step. So, there would be many intermediate models trained for the same prediction. In our base CNN, there are three layers. Consequently, it is possible to have base predictors with 2, 3 and 4 trainable hidden layers. Table \ref{table:deep} draws a comparison between these models.

\begin{table}
	\centering
	\scriptsize
	\begin{tabular}{c| c| c | c | c}
		
		Target stock \textbackslash Model & 2D-CNNpred  & B pred (2 layers)& B pred (3 layers) & B pred (4 layers) \\ 
		\hline \hline
		Whole data & 0.4942 & 0.5022 & 0.5050 & \textbf{0.5111} \\
		XOM & 0.4969 & 0.5079 & 0.5059 & \textbf{0.5109}\\ 
		JPM & 0.4936 & 0.5075 & 0.5017 & \textbf{0.5176} \\
		AAPL & 0.4889 & 0.5004 & \textbf{0.5076} & 0.5058 \\
		MSFT & 0.4961 & 0.5039 & 0.5157 & \textbf{0.5213} \\
		GE & 0.4973 & 0.5095 & 0.4955 & 0.\textbf{5161} \\
		JNJ & 0.4840 & 0.4921 & 0.5021 & \textbf{0.5071} \\
		WFC & 0.4949 & \textbf{0.5104} & 0.5092 & 0.5102 \\
		AMZN & 0.4962 & 0.5068 & 0.5069 & \textbf{0.5114}\\
		
	\end{tabular}
	\caption{F-measure of base predictor with different layers}
	\label{table:deep}
\end{table}

The third experiment is done with the aim of testing robustness of U-CNNpred in forecasting new stocks that have not been used in training. We divide this experiment into two sub experiments of U.S. indices and non-U.S. indices. Another goal of this experiment is to measure the robustness of U-CNNpred in presence of a time gap between training data and test data. In this experiment, training and validation data end in Jul 2015 and test data starts in Apr 2016. Table \ref{table:us_indices} shows the results of applying variations of base predictors that are build in different stages of fine-tuning process on 5 new U.S. indices. We also report the best performance of prediction of CNN-cor and 2D-CNNpred. Furthermore, other baseline algorithms are applied to the 5 mentioned indices. Since train data used in training of base predictor could be related to the 5 U.S indices and bias the results, we also measure the robustness of our framework using indices outside of the U.S. market. So, the same procedure is applied to new dataset that includes 9 renowned stock market indices from around the world. Table \ref{table:other_indices} compares the performance of different algorithms applied on world\textquotesingle s famous indices.

\begin{table}
	\centering
	\scriptsize
	\begin{tabular}{c| c| c| c| c| c | c }
		Target stock \textbackslash Model & Tech+ANN & PCA+ANN & CNN-cor & 2D-CNNpred & P pred & C pred\\  
		\hline \hline
		S\&P 500 & 0.4469 & 0.4237 & 0.3928 & 0.4813 & \textbf{0.4941} & 0.4906 \\
		NASDAQ & 0.4199 & 0.4136 & 0.3769 & 0.4861 & 0.5021 & \textbf{0.5027}\\ 
		Dow 30 & 0.4150 & 0.4283 & 0.39 & 0.4862 & \textbf{0.4924} & 0.4905\\
		NYSE & 0.4071 & 0.4260 & 0.3906 & 0.4785 & 0.4865 & \textbf{0.4871}\\
		RUSSELL 2000 & 0.4525 & 0.4279 & 0.3924 & 0.4849 & \textbf{0.4905} & 0.4904\\
		Average & 0.4283 & 0.4239 & 0.3885 & 0.4834 & \textbf{0.4931} & 0.4923\\
		
	\end{tabular}
\caption{F-measure of algorithms in prediction of U.S. indices}
\label{table:us_indices}
\end{table}

\begin{table}
	\centering
	\scriptsize
	\begin{tabular}{c| c| c| c| c| c | c }
		Target stock \textbackslash Model & Tech+ANN & PCA+ANN & CNN-cor & 2D-CNNpred & P pred & C pred\\  
		\hline \hline				
		DAX & 0.4344 & 0.4455 & 0.3219 & 0.4601 & \textbf{0.4910} & 0.4842 \\
		KOSPI & 0.4699 & \textbf{0.4985} & 0.3180 & 0.4641 & 0.4931 & 0.4938 \\ 
		Nikkei & 0.4183 & 0.4114 & 0.3269 & 0.4620 & \textbf{0.4881} & 0.4881 \\3
		CAC & 0.4766 & 0.4565 & 0.3421 & 0.4610 & \textbf{0.4859} & 0.4842 \\
		FTSE & 0.4371 & 0.4566 & 0.3328 & 0.4491 & \textbf{0.4876} & 0.4850 \\ 
		HSI & 0.4518 & 0.4184 & 0.3156 & 0.4644 & 0.4971 & \textbf{0.4979} \\
		SSE & 0.4520 & 0.4877 & 0.3122 & 0.4753 & \textbf{0.4874} & 0.4866 \\
		BSE 30 & 0.4196 & \textbf{0.5067} & 0.3240 & 0.4709 & 0.4804 & 0.4791\\ 
		NIFTI 50 & 0.4366 & 0.4803 & 0.3274 & 0.4694 & \textbf{0.4861} & 0.4837 \\
		Average & 0.4440 & 0.4624 & 0.3246 & 0.4640 & \textbf{0.4886} & 0.4870 \\
	
	\end{tabular}
	\caption{F-measure of algorithms in prediction of world\textquotesingle s famous indices}
	\label{table:other_indices}
\end{table}

\section{Discussion}
It is obvious from the results that our proposed framework outperforms all of the baseline algorithms in almost all the reported experiments. Wisely selection of features and parameters of the base CNN could be a reason for superiority of that over the other baseline algorithms. However, when base CNN is compared to the base predictor with the same structure, the superiority of the base predictor is clear. The only difference between these two methods lies in their training process. It seems that layer-wise training takes advantage of finding better initial weights for the network using its supervised approach. 

The second experiment showed that making the CNN deeper boosted the ability of the base predictor in prediction. One explanation may be that the deeper models are able to extract more sophisticated features. Consequently, the deeper version of base predictor showed better F-measure. Another interesting observation is that the performance of base CNN is lower than even the shallowest base predictor. In other words, combination of deep neural network and layer-wise training has been necessary to get better results and making the network deeper leads to sub-optimal convergence when using a traditional training approach.

Universal predictor was able to predict 5 U.S. indices with high F-measures. These results suggest that there are common patterns in the behavior of different stock markets and a deep model trained by data from many stock markets, may be able to extract these patterns and use them to predict other stocks. The results of prediction of non-U.S. indices through transfer learning shows the generality of this phenomenon, and is a promising sign for making more efforts toward automatic extraction of the common behavioral patterns for stock markets and building more powerful universal predictors. 

We deliberately put a gap between periods of time that train, validation and test datasets come from. In our experiments, the test data starts almost 9 months after the training and validation data. The results showed high performance of U-CNNpred which leads to the conclusion that U-CNNpred is also a robust predictor regardless of time.

Both partial fine-tuning and complete fine-tuning showed high performance in prediction of indices around the world. However, the average F-measure of partial-fine tuning was slightly higher than complete fine-tuning. It can be inferred that extracted features of base predictor were able to capture the dynamic of stock markets and fine-tuning the last layer rather than the whole network was better due to preserving the extracted features while making base predictor compatible with new market. 

\section{Conclusion}
Prediction of stock markets is one of the challenging interdisciplinary research areas. While scientists have used deep learning algorithms as a new approach in this domain, none of them have tried to use ideas from transfer learning approach for improving the quality of model training. U-CNNpred utilizes a layer-wise supervised training approach to find better initial weights for its CNN structure and performs a fine tuning step afterwards. The suggested approach improved the performance of prediction in our experiments, compared to all of the baseline algorithms, that shows the potential of the suggested approach. The promising results of this experiment suggest to further investigate the possibility of extracting general patterns that explain the behavior of different markets, either for making better predictions or achieving a better understanding about the dynamics of such markets.

\clearpage

\section*{Appendix I. Description of features}
\small
The list of features that were used in this research is represented in Table \ref{table:features}:

\begin{tiny}
	\begin{longtable}[c]{c| c | c |c |c }
		
		\# & Feature & Description & Type & Source / Calculation  \\
		\hline \hline
		\endhead
		1 & Day & which day of week & Primitive & Pandas \\
		2 & Close & Close price & Primitive & Yahoo Finance \\
		3 & Vol & Relative change of volume & Technical Indicator & TA-Lib \\
		4 & MOM-1 & Return of 2 days before & Technical Indicator & TA-Lib \\
		5 & MOM-2 & Return of 3 days before & Technical Indicator & TA-Lib \\
		6 & MOM-3 & Return of 4 days before & Technical Indicator & TA-Lib \\
		7 & ROC-5 & 5 days Rate of Change & Technical Indicator & TA-Lib \\
		8 & ROC-10 & 10 days Rate of Change & Technical Indicator & TA-Lib \\
		9 & ROC-15 & 15 days Rate of Change & Technical Indicator & TA-Lib \\
		10 & ROC-20 & 20 days Rate of Change & Technical Indicator & TA-Lib \\
		11 & EMA-10 & 10 days Exponential Moving Average & Technical Indicator & TA-Lib \\
		12 & EMA-20 & 20 days Exponential Moving Average & Technical Indicator & TA-Lib \\
		13 & EMA-50 & 50 days Exponential Moving Average & Technical Indicator & TA-Lib \\
		14 & EMA-200 & 200 days Exponential Moving Average & Technical Indicator & TA-Lib \\
		15 & DTB4WK & 4-Week Treasury Bill: Secondary Market Rate & Other & FRED \\
		16 & DTB3 & 3-Month Treasury Bill: Secondary Market Rate & Other & FRED \\
		17 & DTB6 & 6-Month Treasury Bill: Secondary Market Rate & Other & FRED \\
		18 & DGS5 & 5-Year Treasury Constant Maturity Rate & Other & FRED \\
		19 & DGS10 & 10-Year Treasury Constant Maturity Rate & Other & FRED \\
		20 & DAAA & Moody\textquotesingle s Seasoned Aaa Corporate Bond Yield & Other & FRED \\
		21 & DBAA & Moody\textquotesingle s Seasoned Baa Corporate Bond Yield & Other & FRED \\
		22 & TE1 & DGS10-DTB4WK & Other & FRED \\
		23 & TE2 & DGS10-DTB3 & Other & FRED \\
		24 & TE3 & DGS10-DTB6 & Other & FRED \\
		25 & TE5 & DTB3-DTB4WK & Other & FRED \\
		26 & TE6 & DTB6-DTB4WK & Other & FRED \\
		27 & DE1 & DBAA-BAAA & Other & FRED \\
		28 & DE2 & DBAA-DGS10 & Other & FRED \\
		29 & DE4 & DBAA-DTB6 & Other & FRED \\
		30 & DE5 & DBAA-DTB3 & Other & FRED \\
		31 & DE6 & DBAA-DTB4WK & Other & FRED \\
		32 & CTB3M & \specialcell{Change in the market yield on U.S. Treasury securities
			at \\ 3-month constant maturity, quoted on investment
			basis} & Other & FRED \\
		33 & CTB6M & \specialcell{Change in the market yield on U.S. Treasury securities
			at \\ 6-month constant maturity, quoted on investment
			basis} & Other & FRED \\
		34 & CTB1Y & \specialcell{Change in the market yield on U.S. Treasury securities
			at \\ 1-year constant maturity, quoted on investment
			basis} & Other & FRED \\
		35 & Oil & Relative change of oil price(WTI), Oklahoma & Commodity & FRED \\
		36 & Oil & Relative change of oil price(Brent) & Commodity & Investing.com \\
		37 & Oil & Relative change of oil price(WTI) & Commodity & Investing.com \\
		38 & Gold & Relative change of gold price (London market) & Commodity & FRED \\
		39 & Gold-F & Relative change of gold price futures & Commodity & Investing.com \\
		40 & XAU-USD & Relative change of gold spot U.S. dollar & Commodity & Investing.com \\
		41 & XAG-USD & Relative change of silver spot U.S. dollar & Commodity & Investing.com \\
		42 & Gas & Relative change of gas price & Commodity & Investing.com \\
		43 & Silver & Relative change of silver price & Commodity & Investing.com \\
		44 & Copper & Relative change of copper future & Commodity & Investing.com \\
		45 & IXIC & Return of NASDAQ Composite index & World Indices & Yahoo Finance \\
		46 & GSPC & Return of S\&P 500 index & World Indices & Yahoo Finance \\
		47 & DJI & Return of Dow Jones Industrial Average & World Indices & Yahoo Finance \\
		48 & NYSE & Return of NY stock exchange index & World Indices & Yahoo Finance \\
		49 & RUSSELL & Return of RUSSELL 2000 index & World Indices & Yahoo Finance \\
		50 & HSI & Return of Hang Seng index & World Indices & Yahoo Finance \\
		51 & SSE & Return of Shang Hai Stock Exchange Composite index & World Indices & Yahoo Finance \\
		52 & FCHI & Return of CAC 40 & World Indices & Yahoo Finance \\
		53 & FTSE & Return of FTSE 100 & World Indices & Yahoo Finance \\
		54 & GDAXI & Return of DAX & World Indices & Yahoo Finance \\
		55 & USD-Y & Relative change in US dollar to Japanese yen exchange rate  & Exchange Rate & Yahoo Finance \\
		56 & USD-GBP & Relative change in US dollar to British pound exchange rate & Exchange Rate & Yahoo Finance \\
		57 & USD-CAD & Relative change in US dollar to Canadian dollar exchange rate& Exchange Rate & Yahoo Finance \\
		58 & USD-CNY & Relative change in US dollar to Chinese yuan exchange rate& Exchange Rate & Yahoo Finance \\
		59 & USD-AUD & Relative change in US dollar to Australian dollar exchange rate& Exchange Rate & Investing.com \\
		60 & USD-NZD & Relative change in US dollar to New Zealand dollar exchange rate& Exchange Rate & Investing.com \\
		61 & USD-CHF & Relative change in US dollar to Swiss franc exchange rate& Exchange Rate & Investing.com \\
		62 & USD-EUR & Relative change in US dollar to Euro exchange rate& Exchange Rate & Investing.com \\
		
		63 & USDX & Relative change in US dollar index & Exchange Rate & Investing.com \\
		
		64 & XOM & Return of Exon Mobil Corporation & U.S. Companies & Yahoo Finance \\ 
		65 & JPM & Return of JPMorgan Chase \& Co. & U.S. Companies & Yahoo Finance\\
		66 & AAPL & Return of Apple Inc. & U.S. Companies & Yahoo Finance\\
		67 & MSFT & Return of Microsoft Corporation & U.S. Companies & Yahoo Finance\\
		68 & GE & Return of General Electric Company & U.S. Companies & Yahoo Finance\\
		69 & JNJ & Return of Johnson \& Johnson & U.S. Companies & Yahoo Finance\\
		70 & WFC & Return of Wells Fargo \& Company & U.S. Companies & Yahoo Finance\\
		71 & AMZN & Return of Amazon.com Inc. & U.S. Companies & Yahoo Finance\\
		
		72 & FCHI-F & Return of CAC 40 Futures & Futures & Investing.com \\
		73 & FTSE-F & Return of FTSE 100 Futures & Futures & Investing.com \\
		74 & GDAXI-F & Return of DAX Futures & Futures& Investing.com \\
		75 & HSI-F & Return of Hang Seng index Futures & Futures & Investing.com \\
		76 & Nikkei-F & Return of Nikkei index Futures & Futures & Investing.com \\
		77 & KOSPI-F & Return of Korean stock exchange Futures & Futures & Investing.com \\
		78 & IXIC-F & Return of NASDAQ Composite index Futures & Futures& Investing.com \\
		79 & DJI-F & Return of Dow Jones Industrial Average Futures & Futures& Investing.com \\
		80 & S\&P-F & Return of S\&P 500 index Futures & Futures& Investing.com \\
		81 & RUSSELL-F & Return of RUSSELL Futures & Futures& Investing.com \\	
		82 & USDX-F & Relative change in US dollar index futures & Exchange Rate & Investing.com \\
		
		\caption{Description of used features}
		\label{table:features}
	\end{longtable}
\end{tiny}


\bibliography{mybibfile}

\end{document}